\documentclass{article}
\usepackage[utf8]{inputenc}
\usepackage{url}
\usepackage{graphicx,algorithm,algcompatible,amsmath, relsize}
\usepackage{hyperref}
\usepackage{subfigure}
\usepackage{algpseudocode}
\usepackage{amsthm,amssymb,inputenc,titling}
\usepackage{hyperref}
\usepackage[T1]{fontenc}

\usepackage{listings}
\usepackage[T1]{fontenc}

\usepackage{blindtext}
\usepackage{xcolor}
\definecolor{codegreen}{rgb}{0,0.6,0}
\definecolor{codegray}{rgb}{0.5,0.5,0.5}
\definecolor{codepurple}{rgb}{0.58,0,0.82}
\definecolor{backcolour}{rgb}{0.95,0.95,0.92}

\lstdefinestyle{mystyle}{
    backgroundcolor=\color{backcolour},
    commentstyle=\color{codegreen},
    keywordstyle=\color{magenta},
    numberstyle=\tiny\color{codegray},
    stringstyle=\color{codepurple},
    basicstyle=\ttfamily\footnotesize,
    breakatwhitespace=false,
    breaklines=true,
    captionpos=b,
    keepspaces=true,
    numbers=left,
    numbersep=5pt,
    showspaces=false,
    showstringspaces=false,
    showtabs=false,
    tabsize=2
}
\usepackage{blindtext}
\title{Multi-Task Triplet Loss for Named Entity Recognition using Supplementary Text}
\author{
Ryan Siskind
\texttt{\{ryan.siskind@target.com\}}\\
Shalin Shah
\texttt{\{shalin.shah@target.com\}}\\
\\
AI Sciences\\
Target Corporation\\
Sunnyvale, CA 94086, USA \\
}
\date{}
\begin{document}
\maketitle
\begin{abstract}
Retail item data contains many different forms of text like the title of an item, the description of an item, item name and reviews. It is of interest to identify the item name in the other forms of text using a named entity tagger. However, the title of an item and its description are syntactically different (but semantically similar) in that the title is not necessarily a well formed sentence while the description is made up of well formed sentences. In this work, we use a triplet loss to contrast the embeddings of the item title with the description to establish a proof of concept. We find that using the triplet loss in a multi-task NER algorithm improves both the precision and recall by a small percentage. While the improvement is small, we think it is a step in the right direction of using various forms of text in a multi-task algorithm. In addition to precision and recall, the multi-task triplet loss method is also found to significantly improve the exact match accuracy i.e. the accuracy of tagging the entire set of tokens in the text with correct tags.
\end{abstract}
\section{Introduction}
Bayesian Personalized Ranking \cite{rendle2012bpr} \cite{shah2021survey} was one of the initial methods that used the triplet loss for personalized recommender systems. The triplet loss \cite{schroff2015facenet} is similar to the contrastive loss \cite{khosla2020supervised} \cite{burges2010ranknet} \cite{gunel2020supervised} but uses a triplet instead of two sets of embeddings. In this work, we use the triplet loss in a multi-task learning method to learn named entity recognition. An item in the catalog of Target has several types of textual data available that in some way describe the item. There is a title, which is a short description of the item, and there is a description which is a long textual representation of the item. Instead of doing named entity recognition by adding the title and the description as separate rows, we use them together through a triplet loss on the embeddings. We find that this improves the results as compared to BERT-base \cite{devlin2018bert}.\\\\
Contrastive loss and the triplet loss are similar in that they both contrast the record under consideration with other records. In a contrastive loss, a row is learned to be similar to another positive row and it is learned to be not similar to another negative row. In a triplet loss, the objective is to maximize the difference between the similarity of the row under consideration with the positive and negative rows respectively.\\\\
The goal in this paper is to use supplementary text (like item descriptions) with the more important item titles in a multi-task triplet loss.
\begin{figure}[ht]
\label{bert-base-loss}
\centering
\includegraphics[scale=0.4]{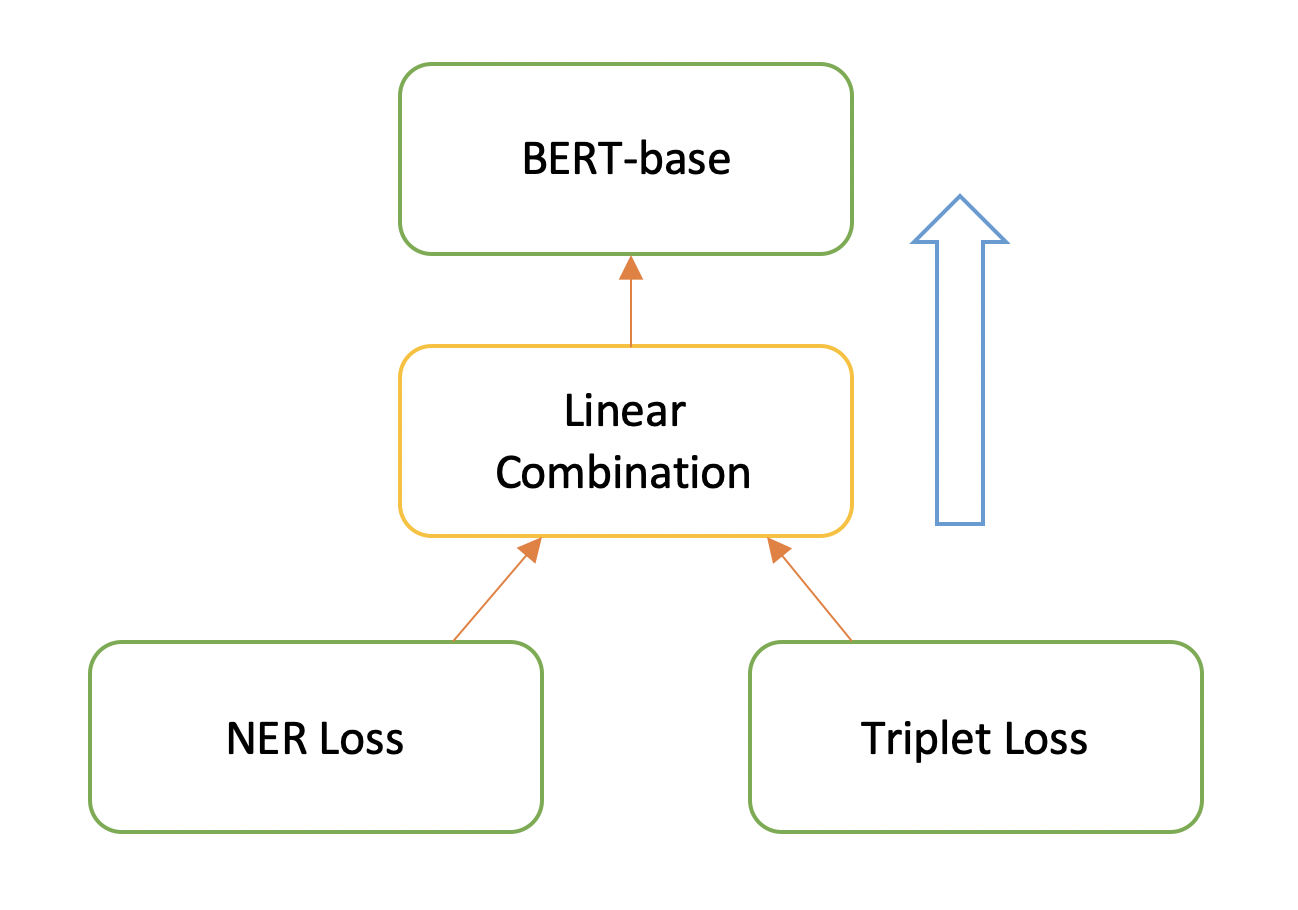}
\caption{Triplet Loss and NER Loss using BERT-base}
\end{figure}
\section{Methods}
We use the triplet loss to contrast the similarities and differences between BERT \cite{devlin2018bert} sentence embeddings of titles and descriptions. This loss is very similar to the loss used in BPR \cite{rendle2012bpr} except that we have two objectives in our loss function and the goal is to learn efficient sentence embeddings as opposed to user and item latent factors.\\\\
Let $t_i$ be the title embedding of the $i^{th}$ title and let $d_p$ and $d_n$ be the sentence embeddings two descriptions, where $d_p$ is the description of the $i^{th}$ item under consideration and $d_n$ is a randomly chosen description of a negative item.\\\\
Then, the triplet loss is the following:\\\\
$c_p = cosine(t_i, d_p)$\\\\
$c_n = cosine(t_i, d_n)$\\\\
$d_i = c_p - c_n$ (where the objective is to maximize)\\\\
$\mathcal{L} = \frac{e^{d_i}}{1 + e^{d_i}}$\\\\
We use this loss $\mathcal{L}$ in a multi-task setting with the named entity recognition loss. We propagate both losses backwards in the neural network. However, during scoring (inference), the descriptions are not used at all.\\\\
We use BERT-base \cite{devlin2018bert} as the common algorithm and we build upon BERT-base through transfer learning. We do not lock any weights during backpropagation.
\begin{figure}[ht]
\label{triplet-loss}
\centering
\includegraphics[scale=0.4]{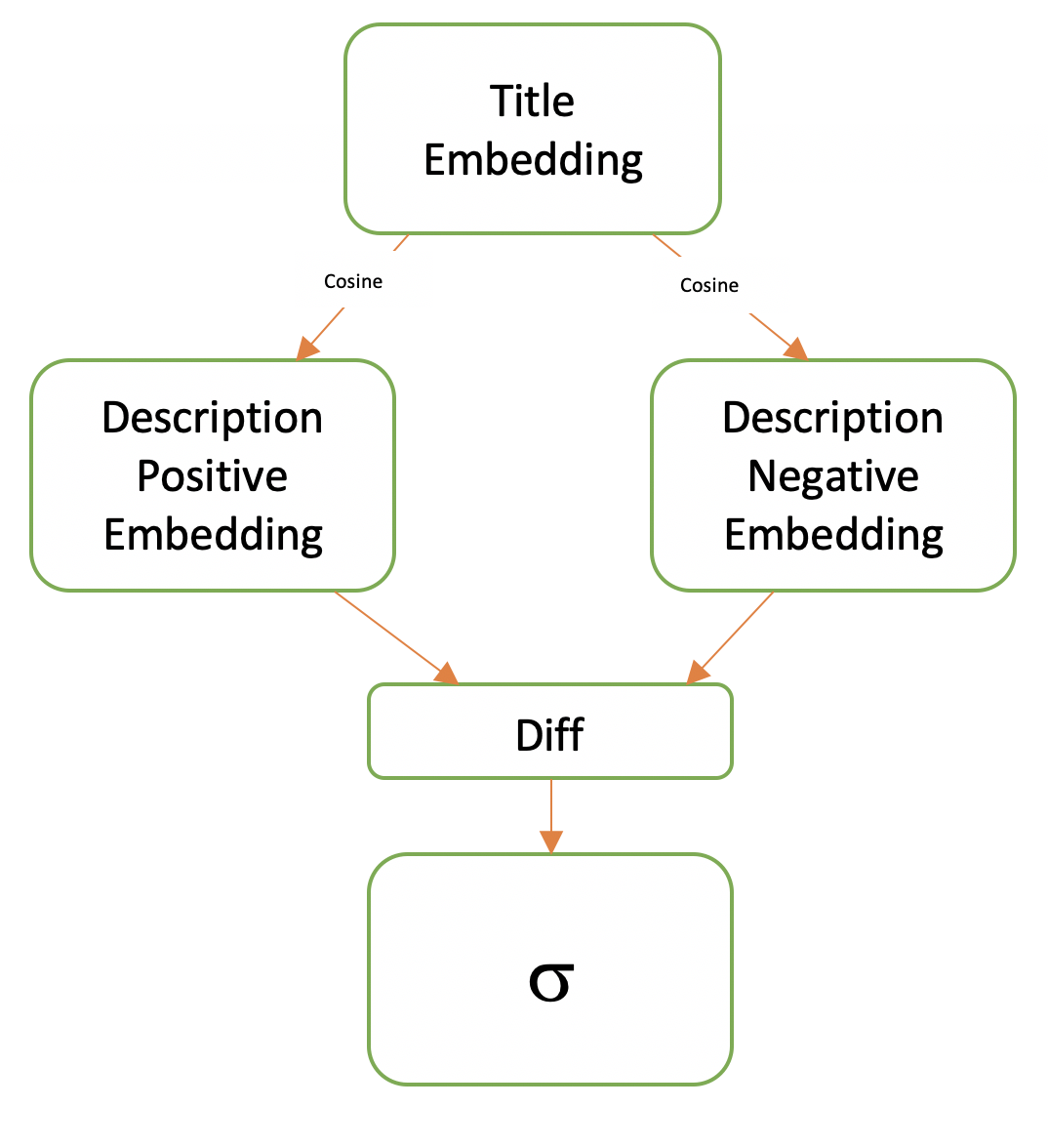}
\caption{Title-Description Triplet Loss}
\end{figure}
\section{Results}
Table 1 shows the results of our BERT-Multitask-Triplet. As the table shows the BERT-Multitask-Triplet algorithm improves upon precision and recall by a percentage point.\\\\
An exact match is a labeling of the sentence in which all tags are correct. As the results show, the BERT-Multitask-Triplet method is able to improve the exact match accuracy by two percentage points which is quite significant.\\\\
The improvement in overall accuracy is marginal, but still relevant to show that the BERT-Multitask-Triplet outperforms BERT-base.
\begin{table}[t]
\caption{Results on a 30\% held out test set (same random seeds)}
\label{results}
\begin{center}
\begin{tabular}{p{3cm}p{1.8cm}p{1.8cm}p{1.8cm}p{1.8cm}}
\hline
\bf Algorithm &\bf Precision &\bf Recall &\bf Exact Matches &\bf Accuracy\\
\hline
BERT-Multitask-Triplet &\bf 78\% &\bf 63\% &\bf 43\% &\bf 85\%\\
\hline
BERT-base &77\% &62\% &41\% &84.7\%\\
\end{tabular}
\end{center}
\end{table}
\section{Conclusion}
In this work, we presented an algorithm BERT-Multitask-Triplet for named entity recognition (NER) and find that after transfer learning, the algorithm is able to outperform BERT-base. Our algorithm is a multi-objective learning algorithm which linearly combines the NER objective with the triplet loss objective. The triplet loss objective uses supplementary text data of the items in the catalog of Target like the item descriptions.\\\\
This method is applicable to many other domains like in medicine where notes of multiple doctors might be available for a patient and the goal is to some form of classification. Instead of concatenating the text, the supplementary text could be used in a multi-objective learning algorithm which contrasts the sentence embeddings using a loss as described above.\\\\
It could be useful to compare algorithms using BERT-large to see if the improvements carry forward as the number of parameters of the neural network increase. It might be useful to compare the algorithm with contrastive losses to see if the triplet loss does indeed work better.
\nocite{*}
\bibliographystyle{unsrt}
\bibliography{ner}

\begin{thebibliography}{1}

\bibitem{rendle2012bpr}
Steffen Rendle, Christoph Freudenthaler, Zeno Gantner, and Lars Schmidt-Thieme.
\newblock Bpr: Bayesian personalized ranking from implicit feedback.
\newblock {\em arXiv preprint arXiv:1205.2618}, 2012.

\bibitem{shah2021survey}
Shalin Shah.
\newblock A survey of latent factor models for recommender systems and
  personalization.
\newblock 2021.

\bibitem{schroff2015facenet}
Florian Schroff, Dmitry Kalenichenko, and James Philbin.
\newblock Facenet: A unified embedding for face recognition and clustering.
\newblock In {\em Proceedings of the IEEE conference on computer vision and
  pattern recognition}, pages 815--823, 2015.

\bibitem{khosla2020supervised}
Prannay Khosla, Piotr Teterwak, Chen Wang, Aaron Sarna, Yonglong Tian, Phillip
  Isola, Aaron Maschinot, Ce~Liu, and Dilip Krishnan.
\newblock Supervised contrastive learning.
\newblock {\em arXiv preprint arXiv:2004.11362}, 2020.

\bibitem{burges2010ranknet}
Christopher~JC Burges.
\newblock From ranknet to lambdarank to lambdamart: An overview.
\newblock {\em Learning}, 11(23-581):81, 2010.

\bibitem{gunel2020supervised}
Beliz Gunel, Jingfei Du, Alexis Conneau, and Ves Stoyanov.
\newblock Supervised contrastive learning for pre-trained language model
  fine-tuning.
\newblock {\em arXiv preprint arXiv:2011.01403}, 2020.

\bibitem{devlin2018bert}
Jacob Devlin, Ming-Wei Chang, Kenton Lee, and Kristina Toutanova.
\newblock Bert: Pre-training of deep bidirectional transformers for language
  understanding.
\newblock {\em arXiv preprint arXiv:1810.04805}, 2018.

\end{thebibliography}
\end{document}